# Unifying Segment Anything in Microscopy with Vision-Language Knowledge

Manyu Li*
Fudan University
24210240029@m.fudan.edu.cn

Ruian He*
Fudan University
rahe16@fudan.edu.cn

Zixian Zhang
Fudan University
23210240062@m.fudan.edu.cn

Chenxi Ma†
Fudan University
cxma17@fudan.edu.cn

Weimin Tan†
Fudan University
wmtan@fudan.edu.cn

Bo Yan†
Fudan University
byan@fudan.edu.cn

## Abstract

*Accurate segmentation of regions of interest in biomedical images holds substantial value in image analysis. Although several foundation models for biomedical segmentation have currently achieved excellent performance on certain datasets, they typically demonstrate sub-optimal performance on unseen domain data. We owe the deficiency to lack of vision-language knowledge before segmentation. Multimodal Large Language Models (MLLMs) bring outstanding understanding and reasoning capabilities to multimodal tasks, which inspires us to leverage MLLMs to inject **V**ision-**L**anguage **K**nowledge (VLK), thereby enabling vision models to demonstrate superior generalization capabilities on cross-domain datasets. In this paper, we propose a novel framework that seamlessly uses MLLMs to guide SAM in learning microscopy cross-domain data, **u**nifying **S**egment **A**nything in **M**icroscopy, named uLLSAM. Specifically, we propose the **V**ision-**L**anguage **S**emantic **A**lignment (VLSA) module, which injects VLK into Segment Anything Model (SAM). We find that after SAM receives global VLK prompts, its performance improves significantly, but there are deficiencies in boundary contour perception. Therefore, we further propose **S**emantic **B**oundary **R**egularization (SBR) to regularize SAM. Our method achieves performance improvements of 11.8% in SA across 9 in-domain microscopy datasets, achieving state-of-the-art performance. Our method also demonstrates improvements of 9.2% in SA across 10 out-of-domain datasets, exhibiting strong generalization capabilities. Code is available at https://github.com/ieellee/uLLSAM.*

## 1. Introduction

The convergence of advanced human imaging techniques and computational technologies has dramatically accelerated the acquisition of microscopic imagery across diverse imaging conditions, application domains, and modalities. This unprecedented rate of data generation has created a significant bottleneck in the scientific workflow, as the limited number of domain experts available cannot analyze these vast datasets at a pace commensurate with their production [41]. Consequently, there exists an urgent need among specialists for sophisticated tools that facilitate high-quality annotation of newly generated data while simultaneously enabling comprehensive description of structural features, intricate details, and underlying mechanisms. Such tools must be designed to align seamlessly with the specific requirements of domain experts, enabling them to extract meaningful insights efficiently and maintain scientific productivity in the face of ever-expanding data repositories [55, 74]. The development of these annotation solutions represents a critical challenge at the intersection of computer vision, nature language processing, and specialized domains.

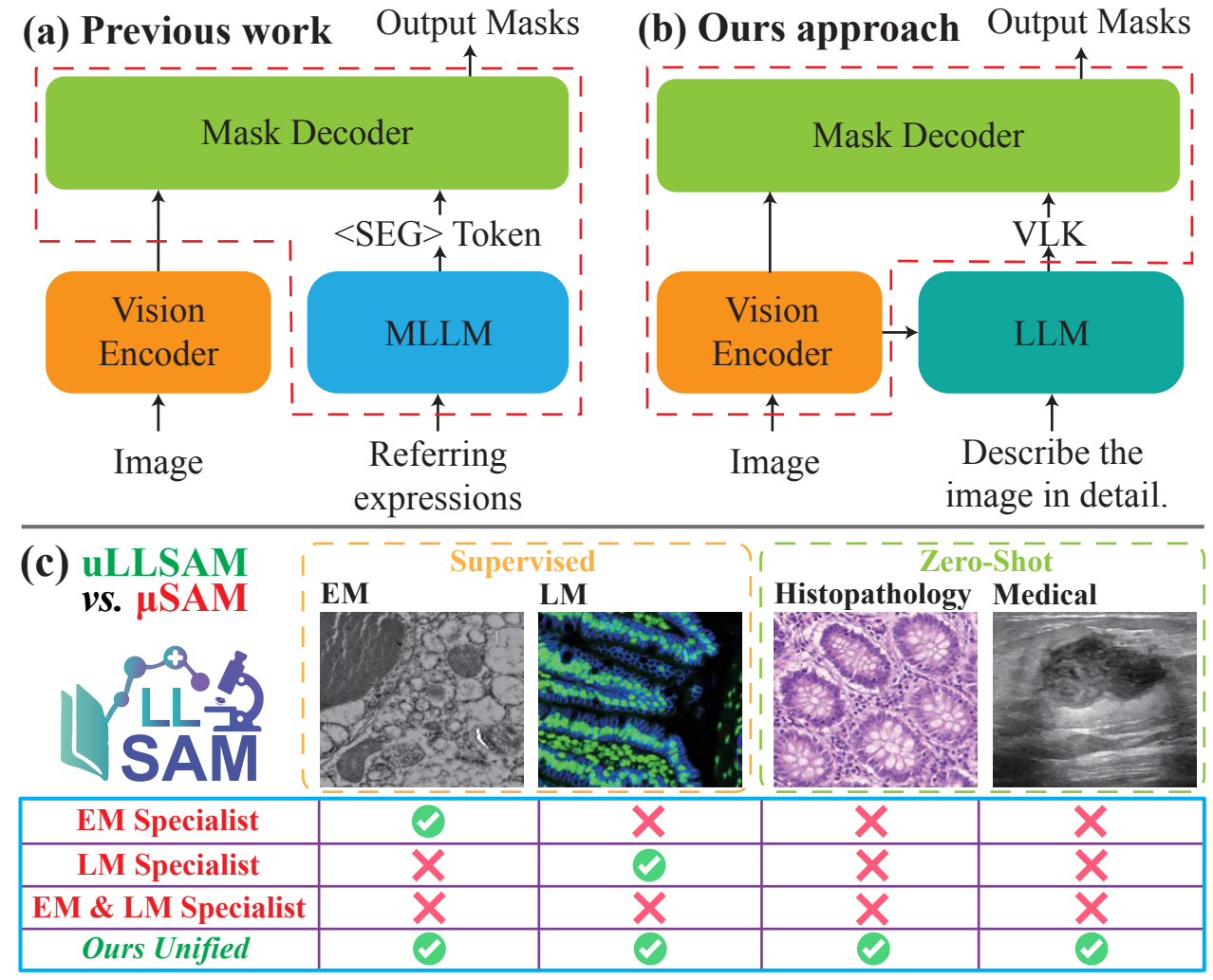

Figure 1. Overview of designs and generalization. (a) Prior work: an **MLLM-centric** SAM+MLLM pipeline. (b) Our approach: a **SAM-centric** SAM+LLM design. (c) Generalization comparison with $\mu$SAM on four structurally similar modalities: EM and LM (supervised), and histopathology and medical (zero-shot).

*These authors contributed equally.
†Corresponding Authors.

To accelerate research for domain scientists in microscopy, numerous foundation models for downstream tasks have been developed, including image restoration [40] and cellular tissue segmentation [3, 42, 44, 51, 64, 76]. Among these, $\mu$SAM [3] has been specifically developed on the foundation of the SAM [28], offering two separate model weights **tailored for** light microscopy (LM) and electron microscopy (EM). These specialized weights enable interactive segmentation, interactive tracking, and fully automated segmentation capabilities. However, these microscopy foundation models exclusively focus on a specific domain, which encounters substantial generalization challenges when deployed across heterogeneous domain data, primarily due to their insufficient integration of vision-language knowledge. Most critically, these models are constructed purely on visual architectures, severely lacking *semantic perception capabilities* when processing data from different domains, a key limitation in the understanding of biological structures.

With the advent of Multimodal Large Language Models (MLLMs) like LLaVA [32] for natural images, numerous works have emerged applying multimodal architectures to downstream visual tasks including referring detection [19], reasoning segmentation [29, 34, 56, 75], visual question answering [26], and visual reasoning [9]. These MLLMs leverage powerful implicit semantic modeling capabilities that mutually enhance feature representation across visual and linguistic components, enabling deeper understanding of image information and different domains. The recent growth of microscopy-centric visual-language datasets [8, 36, 37], particularly BIOMEDICA [37] which collected 24 million high-quality image-text pairs from scientific literature across 12 categories including Microscopy, presents tremendous potential for MLLMs development in the microscopy domain.

As shown in Figure 1 (b), we present uLLSAM. To the best of our knowledge, we are the first framework to explore the integration of MLLMs and SAM in the microscopy domain, aiming to leverage the powerful understanding and reasoning capabilities of MLLMs to inject vision-language knowledge into SAM, thereby enabling SAM to effectively learn cross-domain vision-language knowledge. Specifically, our contributions include:

- **Unified multimodal processing for microscopy data.** We propose uLLSAM, which leverages MLLMs to guide SAM in learning cross-domain vision-language knowledge, achieving improved segmentation performance across different microscopy domains. This approach enables a unified framework for processing both LM and EM data, with significant performance improvements in Figure 2 (a), achieving state-of-the-art results.

Table 1. The key differences from prior methods. Ours is the only approach that jointly leverages SAM and an LLM while allowing LLM-free inference.

| **Methods** | Using SAM? | Using LLM? | Can inference w/o LLM? |
|---|---|---|---|
| LISA [29] | ✓ | ✓ | ✗ |
| GLaMM [56] | ✓ | ✓ | ✗ |
| GSVA [70] | ✓ | ✓ | ✗ |
| Llm-seg [67] | ✓ | ✓ | ✗ |
| EVF-SAM [75] | ✓ | ✓ | ✗ |
| BiomedParse [76] | ✗ | ✗ | - |
| MedSAM [42] | ✓ | ✗ | - |
| $\mu$SAM [3] | ✓ | ✗ | - |
| Ours | ✓ | ✓ | ✓ |

- **Vision-language knowledge injection.** We propose the Visual-Language Semantic Alignment (VLSA) module to inject vision-language knowledge into SAM during training, and during inference the VLK can be omitted (fast mode), trading only a **2.3%** performance drop for a **72.8%** reduction in first-pass inference time. Due to the decreased boundary awareness capability of SAM after incorporating vision-language knowledge, we propose Semantic Boundary Regularization (SBR) to enhance SAM's boundary awareness capability.
- **Strong cross-domain generalization.** uLLSAM demonstrates robust zero-shot generalization capabilities, outperforming existing methods in cross-domain scenarios. It achieves substantial improvements on 10 unseen datasets from EM, LM, pathology, and medical domains, showcasing its ability to adapt to new domains without requiring additional training.
- **Friendly interactive interface.** We follow $\mu$SAM's expert-in-the-loop paradigm and develop a user-friendly interactive GUI, providing domain experts with a powerful tool. Details in appendix Sec 8.

## 2. Related work

### 2.1. Extending SAM with MLLMs

SAM's remarkable generalization on natural images has led to extensions like LISA [29], GLaMM [56], GSVA [70] and LLM-Seg [67]. These methods excel in referring segmentation for natural images but struggle with specialized domains like microscopy due to scarce high-quality data and limited pre-trained weight generalization. EVF-SAM [75] uses early fusion strategies but loses crucial point-level interaction features. The main distinctions between our method and prior methods are summarized in Table 1. Our approach leverages MLLMs to train a generalized image encoder, maintains point-level interaction, and uses the VLSA module to inject VLK into SAM. With a higher-resolution image model and InternLM2.5-1.8B [10], our method offers better flexibility and precision for tasks requiring nuanced understanding of visual and language inputs.

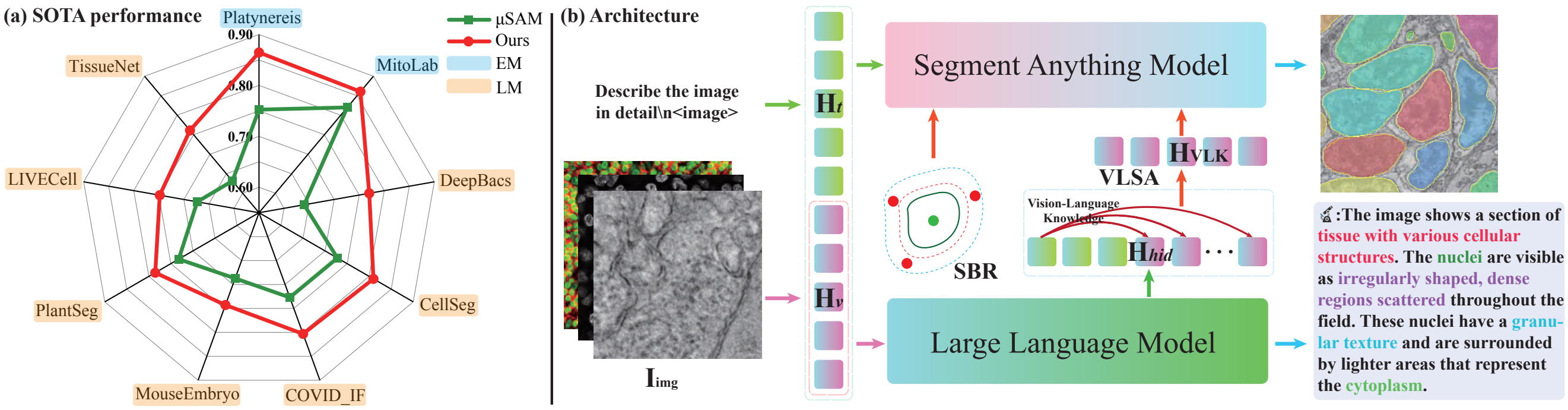

Figure 2. (a) SOTA performance on SA metric. Our method surpasses $\mu$SAM across diverse microscopy segmentation benchmarks, showing consistent gains. (b) Architecture. We propose uLLSAM, where SAM and the LLM share an image encoder. SAM is refined by SBR and injected with vision–language knowledge via VLSA module. At inference, SAM and the LLM can be decoupled, which is a **key distinction** from prior SAM+LLM work.

### 2.2. Interactive segmentation in biomedical fields

Biomedical image segmentation has advanced with models like BiomedParse [76], MedSAM [42], and $\mu$SAM [3]. BiomedParse jointly learns segmentation and detection tasks using large datasets and PubMedBERT [24] but cannot perform multi-instance segmentation. MedSAM works across various medical imaging tasks but struggles with vascular structures and rare imaging domains. $\mu$SAM addresses microscopic image segmentation but lacks consistency across different microscope domains. These models face limited generalization and semantic awareness issues. Our approach uses MLLM's vision-language knowledge to improve SAM model generalization, provides an interactive interface with basic image analysis capabilities, and implements the SBR for robust interactive performance, making it more versatile for biomedical imaging tasks.

### 2.3. Application of MLLMs in biomedical fields

MLLMs have broad medical applications including cancer diagnosis [5, 18, 39, 46, 71, 77], diagnostic report interpretation [72], explainable diagnosis [69], and pathology image analysis [38]. Evaluation methods include expert scoring, BLEU [53], and GPT-4 [1] scoring. In biology, MLLMs exploration remains limited, with works discussing multimodal foundation models in molecular cell biology [14] and the GPT-4-based Omega [58] tool for cell segmentation. Unlike Omega, where segmentation and large model components do not interact, our uLLSAM features interaction between these components. To the best of our knowledge, we are the first to explore SAM with MLLMs in microscopy, offering inspiration despite challenges such as the lack of high-quality biological datasets.

## 3. Method

To address the fundamental constraint of $\mu$SAM that restricts its capability to process domain data exclusively through corresponding domain-specific models, we propose uLLSAM, which can handle data from different domains with a unified model. In Sec 3.1, we introduce the background of SAM and $\mu$SAM, followed by a detailed description of our proposed uLLSAM in Sec 3.2. Sec 3.3 will illustrate training strategies of uLLSAM.

### 3.1. Preliminaries: SAM and $\mu$SAM

SAM [28] is a foundation vision model for segmenting anything in natural images, while $\mu$SAM [3] is developed based on SAM for segmenting anything in microscopy. SAM mainly consists of three parts: (1) An image encoder responsible for feature extraction from images. (2) A prompt encoder that processes user input prompts. (3) A mask decoder that generates predicted masks after receiving encoded image features and prompt features. $\mu$SAM was trained with two sets of parameters on LM and EM datasets, with two branches after the image encoder: (1) The first branch connects directly to a decoder, predicting the foreground of each instance, distances to object centers and boundaries, and then post-processing to obtain results. (2) The second branch consists of SAM's prompt encoder and mask decoder, which generates a positive point in undersegmented regions and a negative point in incorrectly segmented regions to correct the results after each forward pass. More details can be found in [3]. The features of biomedical images vary significantly across different domains [44, 49, 51, 55, 64, 65]. MLLMs can provide powerful multimodal understanding and reasoning capabilities [73], which brings hope for unifying cross-domain biomedical images. Sec 3.2 will elaborate in detail on our method for injecting vision-language knowledge into SAM.

### 3.2. Ours: uLLSAM

Our motivation is illustrated in sub-figure (c) of Figure 1, where $\mu$SAM can only process specific domain data using specific weights, and lacks analytical descriptions of images. uLLSAM requires only one set of model parameters to process multiple domains of microscopy data, and can also handle histopathology and medical domain similar to microscopy. The overall architecture of uLLSAM is shown in Figure 2 (b), where the VLSA module inject vision-language knowledge into SAM's prompt encoder, and the Semantic Boundary Regularization (SBR) strategy is responsible for generating prompt points based on ground truth masks. The specific details will be described in the following subsections.

#### 3.2.1. Vision-language semantic alignment

SAM and LLM share the same Vision Transformer [16] (ViT-B/16). For vision-language alignment, we follow the alignment method in LLaVA [32]. Specifically, we employ a visual projection layer $\mathbf{M_{proj}}$ , with a pixel shuffle [60] function $\mathbf{pix}(\cdot, ratio) : \mathbb{R}^{B\times H\times W\times C} \rightarrow \mathbb{R}^{B\times(H\times ratio)\times(W\times ratio)\times(C/ratio^2)}$ used to adjust the number of visual tokens according to $ratio$. Given an input image $\mathbf{I} \in \mathbb{R}^{H\times W\times 3}$, visual encoder $f_{\theta_{\mathbf{vis}}}(\cdot)$ and LLM decoder $f_{\theta_{\mathbf{llm}}}(\cdot)$, our data flow process is formulated as shown in Eq. 1.

$$\begin{cases} \mathbf{H_{v'}} = \mathbf{pix}(\mathbf{H_v}, 0.5), \text{with } \mathbf{H_v} = f_{\theta_{\mathbf{vis}}}(\mathbf{I}) \\ \mathbf{H}_{hid} = f_{\theta_{\mathbf{llm}}}(\text{concat}(\mathbf{M_{proj}} \times \mathbf{H_{v'}}, \mathbf{H_t})) \\ \mathbf{H_{VLK}} = \text{VLSA}(\mathbf{H}_{hid}) \end{cases} \tag{1}$$

After obtaining the hidden states $\mathbf{H}_{hid}$ from the final layer of the LLM, the VLSA module further processes $\mathbf{H}_{hid}$. Specifically, the VLSA module first separates the visual tokens from $\mathbf{H}_{hid}$, then uses the $\mathbf{pix}(\cdot, ratio)$ operator to adjust the number of visual tokens, and finally employs components such as **layernorm** and **MLP** to modify the dimension of each token so that $\mathbf{H_{VLK}}$ can be injected into SAM's prompt encoder. To ensure numerical stability during training, we additionally introduce scaling factors $\alpha$ and shift factors $\beta$, as shown in Eq. 2.

$$\mathbf{Dense}_{\text{embed}} = \begin{cases} \alpha \times \mathbf{H_{VLK}} + \beta, & \text{Ours} \\ no_mask_embeddings, & \mu\text{SAM} \end{cases} \tag{2}$$

#### 3.2.2. Semantic boundary regularization

During training of uLLSAM, for each instance mask $\mathcal{M}$ we generate $N_p$ positive point and $N_n$ negative points following the Semantic Boundary Regularization (SBR) strategy. Let $\varepsilon$ denote the morphological erosion operator and define the $e$-times eroded interior as

$$\mathcal{E}_e(\mathcal{M}) \triangleq \underbrace{\varepsilon \circ \cdots \circ \varepsilon}_{e \text{ times}}(\mathcal{M}) \tag{3}$$

Positive points are preferentially drawn from this high-confidence interior region. Let $(W, H)$ be the image size, $N_p$ the number of positive points to generate (default $N_p$=1), and let the centroid of $\mathcal{M}$ be

$$\mathbf{c}_{\mathcal{M}} = \left( \tfrac{1}{|\mathcal{M}|} \sum_{p\in\mathcal{M}} x_p,\ \tfrac{1}{|\mathcal{M}|} \sum_{p\in\mathcal{M}} y_p \right) \tag{4}$$

We use $\text{UnifNoRep}(\mathcal{S}, n)$ to denote uniform sampling *without* replacement of $n$ points from $\mathcal{S}$ (valid when $|\mathcal{S}| \geq n$), and $\text{Unif}(\mathcal{S}, n)$ for uniform sampling *with* replacement (valid when $|\mathcal{S}| > 0$), $L_e = |\mathcal{E}_e(\mathcal{M})|$. The positive-point set $\mathcal{P}$ is then

$$\mathcal{P} = \begin{cases} \text{UnifNoRep}\left(\mathcal{E}_e(\mathcal{M}),\ N_p\right), & \text{if } L_e \geq N_p, \\ \text{Unif}\left(\mathcal{E}_e(\mathcal{M}),\ N_p\right), & \text{if } L_e < N_p, \\ \{\mathbf{c}_{\mathcal{M}}\}^{N_p}, & \text{if } |\mathcal{M}| > 0, \\ \{(W/2,\ H/2)\}^{N_p}, & \text{otherwise.} \end{cases} \tag{5}$$

Negative points (three per instance by default) are sampled from low-confidence regions near or outside the boundary (e.g., using a dilated band), we sample background negatives from a thin band outside the object boundary, at a distance from $d_{min}$ to $d_{max}$ pixels ($d_{min} = 9, d_{max} = 11$ by default). Let $d(p, \partial\mathcal{M})$ denote the euclidean distance from pixel $p$ to the boundary $\partial\mathcal{M}$, and let $\Omega$ be the image domain. Define the boundary-adjacent band

$$\mathcal{B} \triangleq \{\, p \in \Omega \setminus \mathcal{M} \mid d_{min} \leq d(p, \partial\mathcal{M}) \leq d_{max} \} \tag{6}$$

To construct a far-background fallback, let $\delta_r(\cdot)$ denote morphological dilation by $r$ pixels and define

$$\mathcal{O} \triangleq \Omega \setminus \left( \delta_{d_{max}}(\mathcal{M}) \cup \mathcal{M} \right) \tag{7}$$

Let $N_n$ be the number of negative points per instance (default $N_n$=3). The negative-point set $\mathcal{N}$ is

$$\mathcal{N} = \begin{cases} \text{UnifNoRep}\left(\mathcal{B},\ N_n\right), & \text{if } |\mathcal{B}| \geq N_n, \\ \text{Unif}\left(\mathcal{B},\ N_n\right), & \text{if } |\mathcal{B}| < N_n, \\ \text{UnifNoRep}\left(\mathcal{O},\ N_n\right), & \text{else if } |\mathcal{O}| \geq N_n, \\ \text{Unif}\left(\mathcal{O},\ N_n\right), & \text{else if } |\mathcal{O}| < N_n, \\ \text{Unif}\left(\Omega \setminus \mathcal{M},\ N_n\right), & \text{otherwise.} \end{cases} \tag{8}$$

The SBR negative sampling strategies provides explicit semantic boundary constraints for training SAM: by placing negatives just outside $\partial\mathcal{M}$, the model learns sharp instance boundaries while remaining robust via far-background fallbacks when the boundary band is sparse.

#### 3.2.3. Fast inference without VLK

As shown in Fig. 1(b), our method is weakly coupled with the LLM, allowing it to be discarded during inference. Further details can be found in Sec. 4.3.

Table 2. Overall performance across nine datasets. **General interactive segmentation models** generalize poorly to the microscopy modality; even the latest supervised SOTA in biology, **CellPoseSAM**, shows limited transfer there. We compare **Specialist** and **Generalist** variants to preliminarily assess the effect of injecting VLK; the **Performance Drop** block quantifies the generalization gap and highlights VLK's benefit. $^{\dagger}$ uses Qwen3-1.7B and $^{\ddagger}$ uses InternLM2-1.8B as the LLM; $^{\text{fast}}$ indicates w/o VLK when inferring. $^{*}$ indicates half the training schedule compared with **Unified Models**. Lower-right annotations denote percentage change: red indicates an increase, green a decrease. Light blue represents EM, orange represents LM. Unless otherwise specified, these notations are assumed consistent hereafter.

| Methods | PY | ML | DB | CS | CI | ME | PS | LC | TN | Avg SA |
|---|---|---|---|---|---|---|---|---|---|---|
| ***General Interactive Segmentation Models*** ↑ | | | | | | | | | | |
| SAM [28] | 4.5 | 19.9 | 4.3 | 3.3 | 2.1 | 9.8 | 8.1 | 2.5 | 4.6 | 6.6 |
| SAM2 [57] | 4.5 | 22.6 | 5.6 | 4.4 | 3.1 | 13.5 | 14.1 | 6.0 | 4.1 | 8.7 |
| SAM-HQ [27] | 4.1 | 17.3 | 3.0 | 3.0 | 1.9 | 7.3 | 5.5 | 1.0 | 0.9 | 4.9 |
| MedSAM [42] | 18.9 | 9.8 | 4.6 | 15.1 | 11.0 | 12.3 | 5.4 | 6.0 | 5.8 | 9.9 |
| ***SOTA Supervised Models*** ↑ | | | | | | | | | | |
| CellposeSAM [50] | 26.1 | 17.2 | 80.6 | 52.3 | 44.4 | 34.1 | 23.1 | 49.6 | 46.8 | 41.6 |
| ***Specialist Models (only trained on one domain dataset)*** ↑ | | | | | | | | | | |
| $\mu$SAM$^{*}$ [3] | 70.3 | 80.7 | 61.9 | 73.2 | 74.2 | 61.5 | 65.8 | 67.3 | 56.3 | 67.9 |
| uLLSAM$^{*}$ | $77.7_{10.5}$ | $83.1_{3.0}$ | $68.3_{10.3}$ | $78.3_{7.0}$ | $78.1_{5.3}$ | $67.9_{10.4}$ | $73.6_{11.9}$ | $72.7_{8.0}$ | $61.8_{9.8}$ | $73.5_{8.2}$ |
| ***Generalist Models (only trained on one domain dataset)*** ↑ | | | | | | | | | | |
| $\mu$SAM$^{*}$ [3] | 59.7 | 61.9 | 42.9 | 50.4 | 52.2 | 41.1 | 40.6 | 31.8 | 47.2 | 47.5 |
| uLLSAM$^{*}$ | $66.4_{11.2}$ | $70.8_{14.4}$ | $50.9_{18.6}$ | $63.7_{26.4}$ | $64.6_{23.8}$ | $50.4_{22.6}$ | $50.3_{23.9}$ | $50.4_{58.5}$ | $58.2_{23.3}$ | $58.4_{22.9}$ |
| ***Performance Drop (Specialist vs Generalist)*** ↓ | | | | | | | | | | |
| $\mu$SAM$^{*}$ [3] | 10.6 | 18.8 | 19.0 | 22.8 | 22.0 | 20.4 | 25.2 | 35.5 | 9.1 | 20.4 |
| uLLSAM$^{*}$ | $11.3_{6.6}$ | $12.3_{34.6}$ | $17.4_{8.4}$ | $14.6_{36.0}$ | $13.5_{38.6}$ | $17.5_{14.2}$ | $23.3_{7.5}$ | $22.3_{37.2}$ | $3.6_{60.4}$ | $15.1_{26.0}$ |
| ***Unified Models (trained on EM + LM datasets)*** ↑ | | | | | | | | | | |
| $\mu$SAM [3] | 75.2 | 82.0 | 64.0 | 72.8 | 72.7 | 68.7 | 73.2 | 67.3 | 63.1 | 71.0 |
| uLLSAM$^{\dagger}$ | $86.6_{15.2}$ | $85.8_{4.6}$ | $78.0_{21.9}$ | $81.0_{11.3}$ | $80.7_{11.0}$ | $73.4_{6.8}$ | $78.3_{7.0}$ | $74.5_{10.7}$ | $76.2_{20.8}$ | $79.4_{11.8}$ |
| uLLSAM$^{\text{fast}}$ | $85.7_{14.0}$ | $85.2_{3.9}$ | $75.8_{18.4}$ | $78.8_{8.2}$ | $77.7_{6.9}$ | $72.0_{4.8}$ | $77.1_{5.3}$ | $71.9_{6.8}$ | $74.3_{17.7}$ | $77.6_{9.3}$ |
| uLLSAM$^{\ddagger}$ | $86.5_{15.0}$ | $86.0_{4.9}$ | $77.0_{20.3}$ | $81.0_{11.3}$ | $80.3_{10.5}$ | $74.3_{8.2}$ | $78.6_{7.4}$ | $74.8_{11.1}$ | $76.1_{20.6}$ | $79.4_{11.8}$ |

### 3.3. Training strategy of uLLSAM

Our uLLSAM adopts a three-stage training approach: vision-language alignment, supervised fine-tuning (SFT), and interactive SAM training. More details (prompt template, etc) can be found in appendix Sec 6.

**Stage 1: Vision-text alignment pretraining.** This stage aligns features from the visual encoder with the language model's feature space through a vision projection layer. We sampled approximately 80K microscopy image-text pairs from the BIOMEDICA [37] dataset.

**Stage 2: supervised fine-tuning.** Due to the scarcity of microscopy datasets with both instance segmentation labels and high-quality text descriptions, we leveraged Qwen2.5VL-72B-Instruct [4] to generate detailed textual descriptions for 9 LM and EM datasets.

**Stage 3: interactive SAM training.** Similar to MedSAM [42] training, we exclusively use point prompts as interactive input, as points flexibly indicate users' regions of interest. For each instance, we generate points by using SBR strategy for training and select a maximum of 4 random instances per image for loss calculation.

## 4. Experiments

### 4.1. Experimental setup

**Datasets.** We sample 20K 2D images from **two** EM datasets: Platynereis (PY) [66], Mitolab (ML) [13]; and 20K from **seven** LM datasets: DeepBacs (DB) [63], Neurips_CellSeg (CS) [43], COVID_IF (CI) [52], MouseEmbryo (ME) [6], PlantSeg (PS) [68], LIVECell (LC) [17], TissueNet (TN) [23], totaling 40K 2D images for model training, and sample 7.8K from the remaining datasets for model performance validation. Specifically, since the datasets contain 3D data and two-channel TissueNet [23], all data are converted to 2D format for processing, and are padded with zero to create square images before being resized to 1024×1024 resolution. Additionally, we prepared **ten** untrained datasets to test the model's zero-shot performance, including **three** LM: CellPose (CP) [49, 65], Omnipose (OP) [15], OrgaSegment (OS) [30]; **three** EM: Uro-Cell (UC) [45], NucMM-M (NM) [7], MitoNet_Benchmark (MB) [13]; **two** histopathology: GLAS (GA) [62], CoNSeP (CN) [21]; and **two** medical: ISIC2018-task1 (IS) [12], BUSI-benign (BU) [2]. In appendix Sec 7, we evaluate uLLSAM on MicroVQA [8].

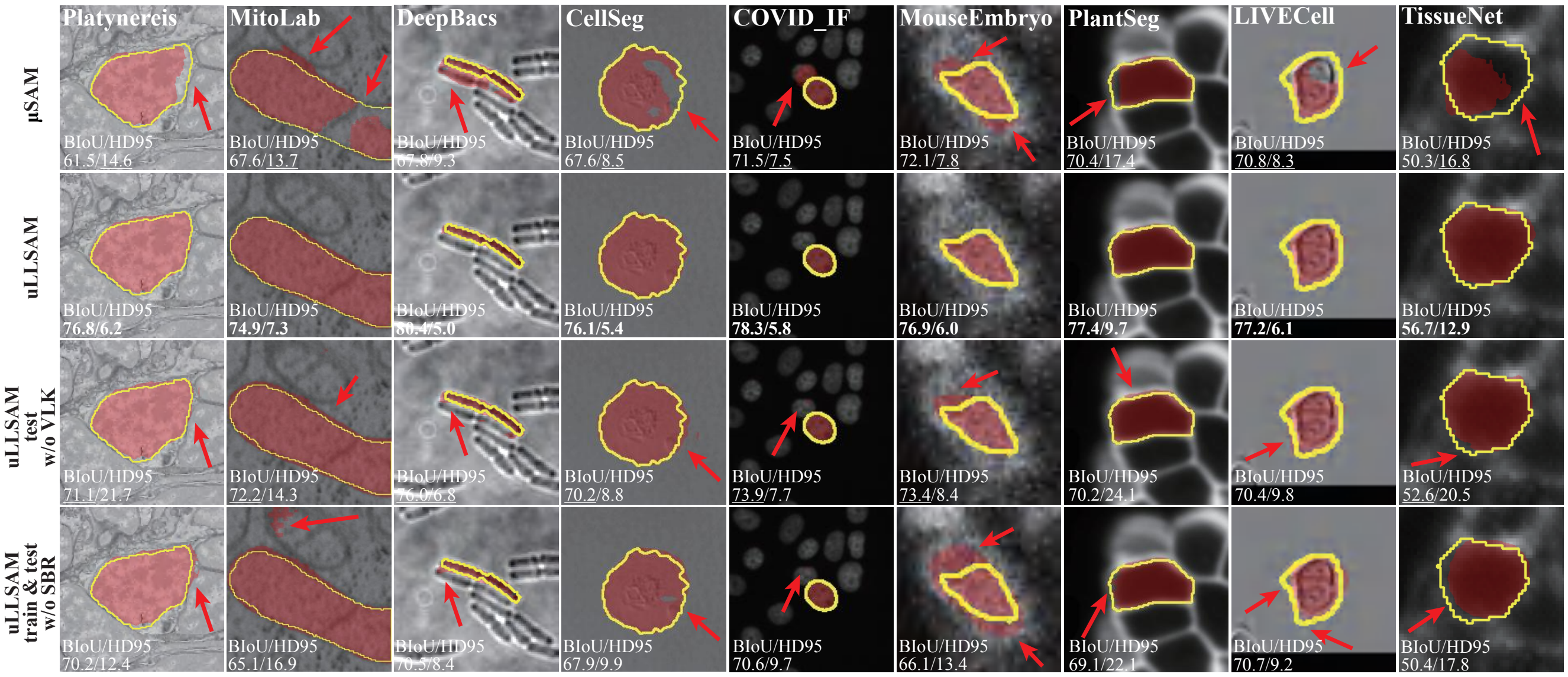

Figure 3. Qualitative evaluation of uLLSAM on the test set. Yellow outlines represent ground truth, with each dataset displaying four images. The first to fourth rows represent $\mu$SAM, uLLSAM, uLLSAM w/o VLK, and uLLSAM w/o SBR, respectively. Note that w/o VLK means VLK is used during training but not during inference. w/o SBR means SBR is not used during either training or inference. The lower-left annotation shows the boundary-granularity evaluation metrics (BIoU and HD95). When VLK is injected but SBR is not used, boundary awareness drops significantly.

**Evaluation metric.** Prompts for the 7.8K validation set are produced via SBR. For evaluation, we adopt the same Segmentation Accuracy (SA) metric as $\mu$SAM [3], using a 0.5 threshold to ensure comparability. We also report the boundary-related metrics BIoU and HD95 in Figure 3, where BIoU is computed over a 13-pixel-wide band.

## 4.2. Comparison experiments

Here we designed three sets of comparison experiments. The first set, referred to as "**General Interactive Segmentation Models**", tested on vanilla SAM and its variants. The second set, referred to as "**Specialist/Generalist Models**", involves training two specialist models (LM-specialist and EM-specialist) with reference to $\mu$SAM, using LM and EM data respectively, and then evaluating the inference performance of these trained specialist models on both in-domain (Specialist) and out-of-domain (Generalist) data. The third set, termed "**Unified Models**", involves combining LM and EM data to train a unified microscopy foundation model, which demonstrated SOTA performance across all datasets. Table 2 shows these qualitative results.

**General interactive segmentation models.** We directly test interactive segmentation performance on the general-purpose foundation vision model SAM and its variants in natural settings. Table 2 (**General Interactive Segmentation Models**) shows the average performance metrics on 9 EM and LM datasets (Dataset abbreviations are detailed in Sec. 4.1.), revealing a significant gap between performance on natural images versus microscopy images. This drives the development of a foundation vision model specifically adapted for the microscopy domain ($\mu$SAM), with requirements for strong generalization capabilities.

**Specialist and generalist models.** However, due to its poor generalization capability, $\mu$SAM exhibits suboptimal performance on cross-modal data. Therefore, we explored whether MLLMs could guide SAM to learn more enriched cross-modal knowledge. Table 2 shows the results of training $\mu$SAM and uLLSAM specialist models separately on single-modal (EM or LM) datasets, then testing them on both EM and LM datasets. In **Performance Drop**, we measure the per-dataset decrease when moving from *Specialist* to *Generalist* models; a smaller drop indicates stronger generalization, thus quantifying the benefit of injecting VLK. Our method demonstrates robust generalization across all datasets except for a slightly weaker performance on the Platynereis dataset compared to $\mu$SAM. With VLK added to SAM, we observe an average 26% relative generalization gain across nine datasets.

These results demonstrate from another perspective that even when SAM is not trained on specific modal data, MLLMs guidance can significantly improve SAM's zero-shot generalization performance. This experiment also inspired our approach to training a unified microscopy SAM segmentation foundation model.

**Unified models.** Inspired by the experimental results in above paragraph, we attempted to use MLLMs to guide

Table 3. Ablations on SBR hyperparameters, we use 1 positive point and 3 negative points.Since setting erosion larger than 10 removes the smallest instances, we cap the erosion parameter at 10.

| $d_{min}, d_{max}$ | $e$ | PY | ML | DB | CS | CI | ME | PS | LC | TN | Avg SA |
|---|---|---|---|---|---|---|---|---|---|---|---|
| 6, 8 | 10 | 84.1 | 82.7 | 71.8 | 73.8 | 71.3 | 62.1 | 73.7 | 67.3 | 70.8 | 73.1 |
| 12, 14 | 10 | 85.6 | 84.5 | 75.2 | 79.0 | 77.8 | 70.4 | 77.3 | 73.1 | 73.4 | 77.4 |
| 9, 11 | 1 | 85.0 | 84.9 | 70.8 | 74.8 | 71.8 | 68.1 | 72.7 | 63.0 | 70.7 | 73.5 |
| 9, 11 | 3 | 85.7 | 85.3 | 72.5 | 76.9 | 74.5 | 70.0 | 75.0 | 67.2 | 72.3 | 75.5 |
| 9, 11 | 5 | 86.1 | 85.8 | 73.5 | 78.9 | 77.1 | 71.9 | 76.6 | 70.7 | 73.8 | 77.2 |
| 9, 11 | 8 | 86.5 | 85.9 | 77.0 | 80.6 | 79.6 | 73.7 | 78.0 | 73.9 | 75.3 | 78.9 |
| 9, 11 | 10 | **86.5** | **86.0** | **77.0** | **80.8** | **80.3** | **74.3** | **78.6** | **74.8** | **76.1** | **79.4** |

Table 4. Ablations on VLSA design, training strategy, and SBR positive/negative point counts. **SS** indicates **s**cale and **s**hift, **Drop** means dropout. **P** and **N** means the number of positive and negative points, respectively.

| SS | Drop | Pretrain | SFT | P | N | Avg SA |
|---|---|---|---|---|---|---|
| ✗ | ✗ | ✓ | ✓ | 1 | 3 | 78.1 |
| ✗ | ✓ | ✓ | ✓ | 1 | 3 | 78.5 |
| ✓ | ✓ | ✓ | ✓ | 1 | 3 | 78.7 |
| ✓ | ✗ | ✗ | ✗ | 1 | 3 | 79.0 |
| ✓ | ✗ | ✓ | ✗ | 1 | 3 | 78.1 |
| ✓ | ✗ | ✓ | ✓ | 1 | 0 | 71.2 |
| ✓ | ✗ | ✓ | ✓ | 3 | 0 | 68.3 |
| ✓ | ✗ | ✓ | ✓ | 3 | 3 | 77.2 |
| ✓ | ✗ | ✓ | ✓ | 5 | 0 | 70.1 |
| ✓ | ✗ | ✓ | ✓ | 5 | 3 | 74.3 |
| ✓ | ✗ | ✓ | ✓ | 1 | 3 | **79.4** |

SAM in combined training across multimodal microscopy datasets, thereby further validating whether MLLMs can help SAM better learn richer domain vision-language knowledge across different domains. As shown in Table 2, uLLSAM‡ demonstrates comprehensive performance improvements (11.8% in average) over $\mu$SAM in SA metrics. Specifically, on the DeepBacs [63] dataset, we observed substantial gains of 20.3%, with the smallest improvements of 4.9% observed on the MitoLab [13] dataset. Figure 3 shows qualitative evaluation of uLLSAM.

**uLLSAM fast mode.** We highlight the model's fast mode, which significantly accelerates inference without sacrificing accuracy, and uncovers several intriguing findings (see Sec. 4.3).

### 4.3. Ablation experiments

The core idea of uLLSAM is to leverage MLLMs to guide SAM in learning rich domain knowledge (*vision-language knowledge injection*), thereby enabling it to process a wider range of domain data. Here, we conducted three ablation experiments centered on MLLM: The first experiment addresses an uncertainty: since our model introduces additional parameters, it remains unclear whether performance improvements stem from these extra parameters or from SAM genuinely learning richer domain knowledge. Therefore, we attempted to directly remove the **Vision-Language Knowledge** from uLLSAM (fast mode) for performance testing to verify the reason for improvement. The second experiment concerns the design of the VLSA module. The third experiment examines the effectiveness of the SBR and its hyperparameter settings. We also performed additional ablation experiments on the training strategy for SAM.

**Vision-language knowledge injection** We conducted tests on 9 in-domain and 10 out-of-domain datasets, using only the trained SAM component of uLLSAM for inference. Tables 2, 5 shows the performance on in-domain datasets. It can be observed that even without VLK during inference (fast mode), the performance comprehensively surpasses $\mu$SAM. Specifically, the DeepBacs [63] dataset achieved the largest performance improvements in SA, with gains of 18.4%. The MitoLab [13] dataset showed the smallest performance improvements, with gains of 3.9%. The average performance improvement across all datasets was 9.3%. Analysis of the results indicates that even without relying on LLM guidance, the well trained uLLSAM still demonstrates significant performance improvements, which strongly proves that our performance gain is not entirely due to the increase in parameter count. Compared to the complete uLLSAM, using only the SAM component resulted in just 2.3% performance degradation.

Table 5 shows our performance results on 10 out-of-domain datasets. Comparing $\mu$SAM with uLLSAM without the LLM component, the GLAS [62] dataset achieved the highest SA performance improvements of 24.6%. On the CoNSeP [21] dataset, there was a slight performance decrease of 1.7%, with an overall average performance improvement of 4.2%. Even in out-of-domain areas, the generalization ability of uLLSAM using only the SAM component still surpasses $\mu$SAM. This further confirms that MLLMs can guide SAM to learn better multimodal features through vision-language knowledge injection.

**VLSA module** We experimented with different designs of the VLSA model. Due to the gap between vision semantic prompts from MLLMs and SAM's prompt space, we explored the impact on model performance of directly inputting these into the SAM prompt encoder versus using scale and shift factors. We also added a dropout layer to VLSA to investigate whether uLLSAM exhibits overfitting phenomena. Analysis from Table 4 reveals that using learnable scale and shift factors improves model performance, while adding dropout layers actually decreases performance, indicating our model does not suffer from significant overfitting issues.

Table 5. Zero-shot performance on ten unseen datasets. Since CellPoseSAM was trained on datasets such as Cellpose and Omnipose, it attains competitive results on those benchmarks. Notation follows Table 1. Bold mark denotes the **best** performance, and single underlining denotes the second-best performance. Purple represents pathology, and brown represents medical datasets.

| Methods | CP | OP | OS | UC | NM | MB | GA | CN | IS | BU | Avg SA |
|---|---|---|---|---|---|---|---|---|---|---|---|
| SAM [28] | 1.6 | 7.9 | 1.9 | 18.9 | 6.9 | 5.9 | 7.9 | 15.2 | 54.6 | 52.3 | 17.3 |
| SAM-HQ [27] | 1.2 | 7.3 | 2.0 | 19.2 | 2.6 | 4.6 | 5.1 | 10.8 | 42.5 | 46.8 | 14.2 |
| SAM2 [57] | 1.3 | 5.0 | 2.0 | 19.1 | 10.7 | 7.2 | 8.6 | 12.1 | 63.8 | 59.8 | 19.0 |
| MedSAM [42] | 21.5 | 10.6 | 4.1 | 15.5 | 3.4 | 3.7 | 27.5 | 13.1 | **83.2** | 49.0 | 23.2 |
| CellposeSAM [50] | **78.1** | **66.5** | 79.6 | 51.8 | 10.9 | 12.3 | 19.2 | 50.7 | 8.6 | 9.3 | 38.7 |
| $\mu$SAM [3] | 64.3 | 51.8 | 77.2 | 85.0 | 76.5 | 63.5 | 58.1 | 70.4 | 57.4 | 66.4 | 67.1 |
| uLLSAM $^{\dagger}$ | 73.3 | 59.4 | 80.4 | 86.9 | 76.9 | 66.9 | 72.0 | 71.8 | 61.3 | **76.0** | 72.5$_{8.0}$ |
| uLLSAM $^{fast}$ | 71.3 | 55.7 | 78.3 | 86.4 | 76.2 | 64.2 | 69.2 | 69.2 | 59.1 | 69.0 | 69.9$_{4.2}$ |
| uLLSAM $^{\ddagger}$ | 74.0 | 60.7 | **80.9** | **87.4** | **78.3** | **68.2** | **72.4** | **72.2** | 64.1 | 74.4 | **73.3**$_{9.2}$ |

**SBR strategy** The last row of Figure 3 with *uLLSAM w/o SBR* demonstrates that directly injecting VLK causes the model to generate blurred object boundaries, The area indicated by the red arrow represents regions with *over-segmentation*, *under-segmentation* and *inaccurate segmentation*. Analysis from Table 4 shows that SBR brings an average performance improvement of 11.5% in SA, thus confirming the effectiveness of the SBR strategy. As shown in Table 3, we ablate the SBR hyperparameters $d_{min}, d_{max}, e$. The initial values are derived from $\mu$SAM bad cases, with details provided in the supplementary.

**Training strategy** Our uLLSAM is the result of a three-stage training process. Here we explore the impact of each stage on model performance. From Table 4, these results suggest that pretraining and SFT enhance the MLLM's perception in the microscopy domain, thereby providing richer vision–language knowledge.

## 4.4. Zero-shot generalization

**Zero-shot performance on ten additional datasets.** To further verify our model's zero-shot performance and generalizability on cross-modal datasets, we additionally selected 3 LM, 3 EM, 2 histopathology, and 2 medical datasets that were not used during training for further validation. Table 5 shows our experimental results, where our method comprehensively outperforms $\mu$SAM. Specifically, GLAS [62] achieved the largest performance improvements on SA with gains of 24.6%, while the MitoNet_Benchmark [13] showed the smallest improvements of 2.4%. Across all 10 datasets, our method achieved an average performance improvement of 9.2%.

Table 6. Compute cost on a single RTX 3090 GPU ($t_{sam}/t_{mllm}$). $^-$ indicates that replace SBR with random point generation.

| Methods | GPU Mem | First-pass | Sub-passes |
|---|---|---|---|
| $\mu$SAM | 11 GB | 0.31s/- | 0.08s/- |
| uLLSAM$^-$ | 24 GB | 0.30s/0.79s | 0.08s/0.01s |
| uLLSAM$^{fast}$ | 11 GB | 0.31s/- | 0.08s/- |
| uLLSAM | 24 GB | 0.31s/0.79s | 0.08s/0.01s |

**SBR strategy enhances generalization.** Interactive prompt point generation strategies typically influence the quality of segmentation masks. For example, in SAM-HQ [27], TinySAM [61], XraySAM [20], using more diverse positive and negative sample points generally produces higher quality results, though this improvement eventually reaches a plateau. Here we explore how different quantities of positive and negative prompt points affect our model's performance. As shown in Table 4, the model achieves optimal average performance on the dataset when using 1 positive point and 3 negative points, indicating that users generally need to provide only four interactive prompt points to obtain satisfactory baseline results. The 3 negative points significantly determine the object's boundary range, enabling the model to segment with greater confidence.

## 4.5. Computational overheads

Table 6 summarizes the memory consumption and latency on one RTX 3090. *GPU Mem* indicates the minimum VRAM needed to train and to run inference. The two time columns give the inference latency for SAM and MLLM, respectively. uLLSAM's first pass per image is the most expensive; sub-passes amortize cost by reusing image embeddings and VLK. Adding SBR incurs only a very small overhead. The uLLSAM fast mode incurs the same overhead as $\mu$SAM.

## 5. Conclusion

In this paper, we propose uLLSAM, the first foundational model that explores interactive segmentation with MLLMs in the field of microscopy. uLLSAM unifies the processing of light and electron microscopy data, and also demonstrates significant improvements in generalization across cross-domain data. Additionally, our model possesses the capability for microscopic image analysis, which previous foundational models lack (user-friendly interactive GUI available). Moreover, *we find that injecting VLK during SAM training substantially improves generalization, and this improvement is not strongly coupled to the LLM at inference time*. We believe that uLLSAM will greatly accelerate MLLMs research in the microscopy domain and provide valuable insights for related fields. Limitations and future work available in appendix Sec 9.

# Unifying Segment Anything in Microscopy with Vision-Language Knowledge

## Supplementary Material

## 6. Additional training details

**Stage 1: Vision-text alignment pretraining.** During this stage, we trained only the Vision Projection Layer for 6 epochs on four RTX3090 GPUs with a batch size of 3, using AdamW[35] optimizer with a learning rate of 1e-4 and CrossEntropy loss function. Unless specified otherwise, subsequent parameters remain consistent.

**Stage 2: supervised fine-tuning.** This stage aims to enhance our model's semantic understanding capabilities. We trained the Vision Projection Layer and LLM for 2 epochs on a single RTX3090 GPU with a batch size of 1 and a learning rate of 1e-6. The prompt template for Qwen2.5-VL-72B is shown as Figure 4.

**Stage 3: interactive SAM training.** This stage uses a learning rate of 1e-3 for training over 24 epochs, with a batch size of 1 and gradient accumulation steps set to 8 to simulate a larger batch size. For each image, the sam_max_point_bs parameter is set to 4, which means that only a maximum of 4 randomly selected instances per image are used for loss calculation and backpropagation. Training was conducted using 4 RTX 3090 GPUs, with a total training time of approximately 40 hours.

## 7. MicroVQA benchmark

Table 7. Answer accuracy performance of uLLSAM and its base model on the challenging MicroVQA dataset. V, H, and E represent different types of perception tasks. This dataset reflects, to some extent, the model's capability in microscope-based fundamental mechanism analysis. Symbol * represents result borrowed from MicroVQA benchmark.

| Model | Overall | V | H | E |
|---|---|---|---|---|
| Random* | 22.0 | 21.9 | 21.8 | 21.9 |
| Llama-3.2-11b*[22] | 30.3 | 32.4 | 29.3 | 28.7 |
| LLaVA-Mistral-7B*[33] | 39.8 | 31.6 | 43.1 | 37.1 |
| Human* | 50.3 | 52.7 | 47.5 | 51.4 |
| o1*[47] | **52.8** | **55.4** | **50.2** | **53.0** |
| InternVL2.5-2B[10] | 35.6 | 35.1 | 33.6 | 40.0 |
| uLLSAM | **39.0** | **39.2** | **36.1** | **43.8** |

Currently, our model focuses on how to improve the visual general generalization ability of the model, therefore the quality of textual description output and hallucination control are not the focus of our method. However, we attempted to preliminarily explore the reasoning and understanding capabilities of the LLM component through evaluation on a microscopy vision-language reasoning benchmark.

The MicroSAM data set is divided into three categories: expert visual understanding (V), hypothesis generation (H), and experimental proposal (E) based on varying scientific requirements and difficulty levels of the task. We benchmarked uLLSAM against its base model (InternVL2.5-2B) on the MicroVQA dataset, demonstrating a substantial improvement of 9.55% in average accuracy. However, since the parameter count of uLLSAM's MLLM component is significantly smaller than that of o1 [47], there remains a considerable performance gap. Future work could explore methods to enhance uLLSAM's image reasoning capabilities.

## 8. User-friendly interface

To facilitate domain experts' use of our model, we developed a user-friendly graphical interface. The overall interface is shown in Figure 5. Basic operations include: 1. Upload images on the left side, supporting formats such as jpeg, png, tif, etc. 2. Select the model to be loaded. 3. Choose positive or negative points, add prompt points directly by clicking on the image. 4. Click Generate Mask to produce segmentation results. 5. Display segmentation results on the right side. After generating a satisfactory mask, click the Save Instance button to save the instance. Each instance is numbered starting from 1.

## 9. Discussion

**Versatility.** Our method is simple and efficient. For professionals in the computer industry, components such as LLMs and image encoders can be easily replaced to match their computational resource capabilities. For researchers in the biomedical field, we provide a user-friendly interactive interface with extremely low deployment and fine-tuning costs—requiring only a single RTX 3090 GPU for smooth operation.

**Impact of LLM choice.** We directly selected InternLM2.5-1.8B and Qwen3-1.7B as the LLM component of our MLLM, while the visual encoder part was initialized with $\mu$SAM pre-trained weights. Due to computational resource constraints, we did not conduct tests on larger LLMs or different types of LLMs; however, we believe that even with different LLMs[11, 22, 31],

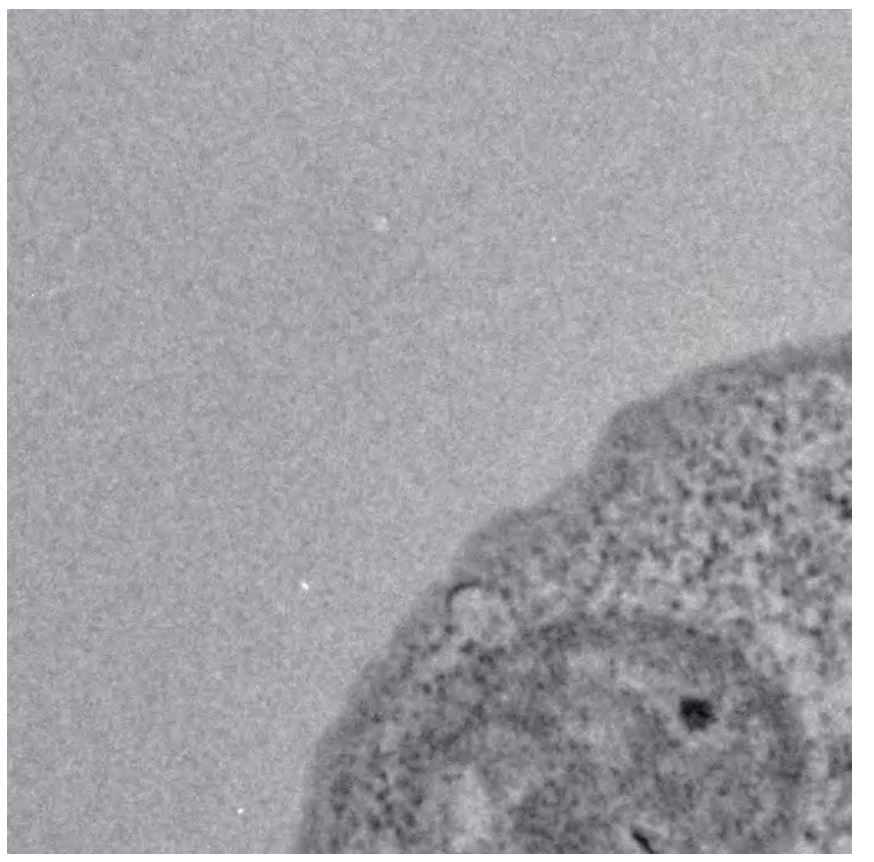

Create detailed captions describing the contents of the given image. Include the object types, textures, double-checking relative positions between objects, etc. Instead of describing the imaginary content, only describing the content one can determine confidently from the image. Do not describe the contents by itemizing them in list form. Minimize aesthetic descriptions as much as possible. Just need to respond with a piece of plain text, as concise as possible. Do not use 'likely', 'perhaps', etc.

The image shows a section of an electron microscopy micrograph. A large, irregularly shaped structure occupies the lower right portion of the frame. This structure has a granular texture with varying shades of gray, indicating density differences within it. The upper left area is mostly empty, showing a uniform background with a fine grainy texture. There are no distinct objects or features in this background region. The boundary between the structure and the background is clear, with the structure extending from the bottom edge upwards and to the right.

Figure 4. When prompting the Qwen2.5-VL-72B model, we primarily focus on object types, texture features, relative positions. We also ensure that the model outputs confident content as much as possible to mitigate hallucinations.

comparable performance can be achieved. In the future, we will further explore the impact of LLM types and parameter sizes on uLLSAM performance.

**Limitations.** Although our method achieves good generalization, there are still shortcomings in terms of interaction methods, text utilization, and other aspects. The **first** limitation lies in the fact that we only consider a single mode of interaction during training. Future work could explore whether diversified prompt interactions may further enhance model generalization. The **second** limitation is that we rely solely on the strong semantic perception capability of LLMs to improve the generalization of SAM, which allows decoupling during inference. However, tasks such as text-guided referring segmentation have not yet been explored, partly due to the lack of expert-level, high-quality annotated data. The **third** limitation is the restriction imposed by computational resources. we have not been able to verify whether larger-scale LLMs could further improve the model's generalization and microscopic image analysis capabilities. One feasible approach is to adopt Parameter-Efficient Fine-Tuning (PEFT) strategies such as LoRA[25]. The **fourth** limitation lies in the fact that we currently only consider a unidirectional interaction between the LLM and SAM. In the future, we will continue to explore how to enable bidirectional interaction between these two components to achieve mutually beneficial outcomes. The fifth limitation is that we currently do not have control interventions for image-level description outputs. In the future, we can explore some reinforcement learning methods [48, 54, 59] to further optimize the model's textual description outputs.

**Broad impact.** To the best of our knowledge, we are the first to explore the application of MLLMs in the field of microscopy, paving the way for future MLLMs research in related areas. Our method can be easily transferred to various scenarios, such as interactive medical image segmentation. And the visual encoder with strong generalization capabilities can be applied to a wide range of downstream tasks. However, the text output by the model currently lacks interpretability and exhibits certain hallucination issues, which may result in the generation of erroneous content. In our future work, we will focus on addressing and optimizing these challenges. We hope our approach will accelerate the progress of MLLMs research in the biomedical domain.

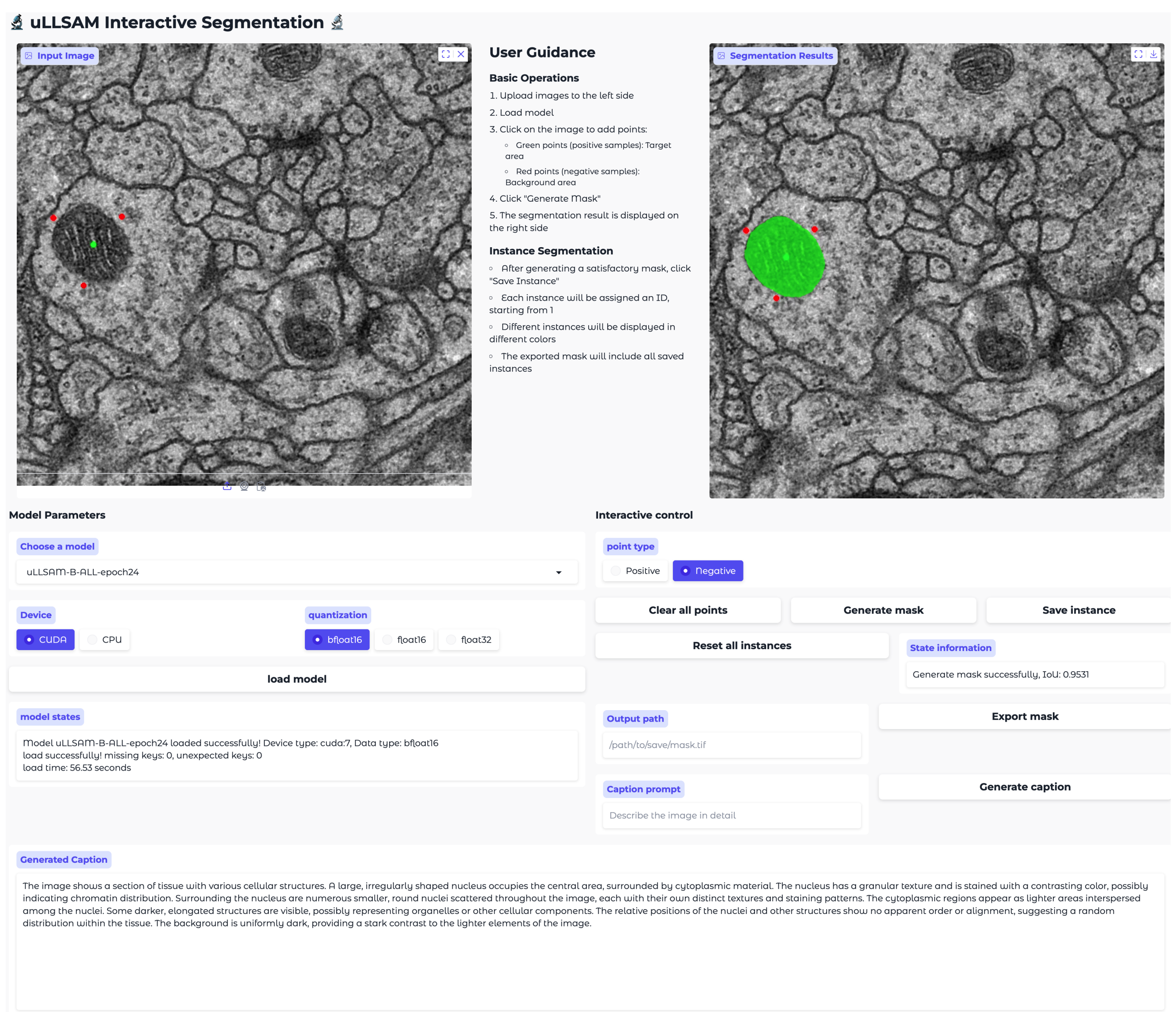

Figure 5. Overall of our user-friendly interface.